\documentclass[sigconf,natbib=true]{acmart}
\AtBeginDocument{%
  \providecommand\BibTeX{{%
    \normalfont B\kern-0.5em{\scshape i\kern-0.25em b}\kern-0.8em\TeX}}}

\setcopyright{acmcopyright}
\copyrightyear{2018}
\acmYear{2018}
\acmDOI{XXXXXXX.XXXXXXX}

\acmConference[Conference acronym 'XX]{Make sure to enter the correct
  conference title from your rights confirmation emai}{June 03--05,
  2018}{Woodstock, NY}
\acmPrice{15.00}
\acmISBN{978-1-4503-XXXX-X/18/06}

\usepackage{times}
\usepackage{latexsym}

\usepackage{graphicx}
\usepackage{float}
\usepackage{subfigure}
\usepackage{amsmath}
\usepackage{multirow}
\usepackage{booktabs}
\usepackage[ruled]{algorithm2e}
\usepackage{bm}
\usepackage{xspace}
\usepackage{booktabs}
\usepackage{enumitem}
\usepackage{CJKutf8}
\newcommand{\songti}[1]{\begin{CJK*}{UTF8}{gbsn} #1  \end{CJK*}}
\usepackage[T1]{fontenc}
\usepackage[utf8]{inputenc}
\usepackage{microtype}
\newcommand{\tabincell}[2]{\begin{tabular}{@{}#1@{}}#2\end{tabular}}

\usepackage{tabularx}
\makeatletter
\def\hlinewd#1{%
\noalign{\ifnum0=`}\fi\hrule \@height #1 %
\futurelet\reserved@a\@xhline}
\makeatother

\begin{document}

\title{Exploring Dense Retrieval for Dialogue Response Selection}


\author{
Tian Lan$^1$, Deng Cai$^2$, Yan Wang, Yixuan Su$^3$, Heyan Huang$^1$, Xian-Ling Mao$^1$ \\
$^1$Beijing Institute of Technology \\
$^2$The Chinese University of Hong Kong \\
$^3$Language Technology Lab, University of Cambridge \\
\texttt{\{lantiangmftby,thisisjcykcd,yanwang.branden\}@gmail.com}\\
\texttt{\{maoxl,hhy63\}@bit.edu.cn} \\
\texttt{ys484@cam.ac.uk}
}
\renewcommand{\shortauthors}{Lan et al.}

\begin{abstract}

Recent progress in deep learning has continuously improved the accuracy of dialogue response selection. In particular, sophisticated neural network architectures are leveraged to capture the rich interactions between dialogue context and response candidates. While remarkably effective, these models also bring in a steep increase in computational cost. Consequently, such models can only be used as a re-rank module in practice. In this study, we present a solution to directly select proper responses from a large corpus or even a nonparallel corpus that only consists of unpaired sentences, using a dense retrieval model. To push the limits of dense retrieval, we design an interaction layer upon the dense retrieval models and apply a set of tailor-designed learning strategies. Our model shows superiority over strong baselines on the conventional re-rank evaluation setting, which is remarkable given its efficiency. To verify the effectiveness of our approach in realistic scenarios, we also conduct full-rank evaluation, where the target is to select proper responses from a full candidate pool that may contain millions of candidates and evaluate them fairly through human annotations. Our proposed model notably outperforms pipeline baselines that integrate fast recall and expressive re-rank modules. Human evaluation results show that enlarging the candidate pool with nonparallel corpora improves response quality further.
\footnote{All our source codes, datasets, model parameters, and other related resources have been publicly available.}

\end{abstract}




\maketitle


\section{Introduction}
Recently, dialogue response selection has attracted increasing attention. The task is to select proper responses given a dialogue context (often a multi-turn conversation) and a set of response candidates.

One common and efficient way to retrieve a response for a given dialogue context is using bag-of-words retrieval functions (e.g., BM25 and docTTTTTquery \cite{Robertson2009ThePR,Cheriton2019FromDT}).
Since only overlapping lexical information is leveraged, it usually fails to capture the semantic relationship between contexts and responses.
Recently, dual-encoder models, also known as \textit{dense retrieval}, have shown strong capability to search more appropriate responses than the former methods. Some typical examples include DPR \cite{Karpukhin2020DensePR}, ColBERT \cite{Zhang2020DCBERTDQ}, and RocketQA \cite{Ren2021RocketQAv2AJ}.
Due to the non-interaction architecture in dual-encoder models, the candidate representations could be pre-computed, supporting the efficient search.
However, precisely because of this architecture, dual-encoder models still underperform the state-of-the-art cross-encoder models in the dialogue response selection task.
The cross-encoder models \cite{Gu2020SpeakerAwareBF,Whang2021DoRS,Xu2021LearningAE,Su2021DialogueRS,han-etal-2021-fine} jointly encode the dialogue context and response to capture rich and fine-grained interactions. But, in inference, these cross-encoder models must compute the matching degree for every possible combination of the context-response pair, which is infeasible to run over millions of candidates in practice. Consequently, they can only be used as a re-rank module in a recall-then-rerank pipeline framework, that is, ranking a small number of candidates recalled from the full candidate pool by a recall module (e.g., BM25 and docTTTTTquery).

In this study, we propose an effective and efficient \textbf{D}ense \textbf{R}etrieval solution for dialogue response selection, named DR-BERT.
Concretely, DR-BERT contains two main components: 
(1) dual-encoder model calculates the context-response matching degree via the inner product of the decoupled dense representations, which supports an efficient search over a huge corpus;
(2) interaction layer ranks the whole candidate set retrieved by dual-encoder models, which could capture the rich information of context-response and response-response interaction in a cheap way.
In training, we optimize them simultaneously with a multi-task learning objective.
Moreover, to push the limits of dense retrieval for dialogue response selection, we employ three simple yet effective learning strategies to train the DR-BERT model, which helps DR-BERT achieve state-of-the-art performance in dialogue response selection. Specifically, 
(1) in-batch contrastive learning effectively optimizes the shared semantic space of conversation context and response;
(2) nonparallel domain adaptive pretraining warms up two BERT models in DR-BERT on nonparallel corpus;
(3) data augmentation supplements a large number of high-quality training samples by leveraging the positive instance expansion and hard negative selection.

Our proposed DR-BERT has the following advantages:
(1) due to the decoupled context and response encoders in the dual-encoder model, the representations of all candidates in the corpus can be pre-computed, indexed, and searched efficiently using off-the-shelf vector search toolkits \cite{JDH17,Karpukhin2020DensePR};
(2) the proposed interaction layer improves the matching degrees further but only introduces a negligible computational overhead by re-using the context embedding and pre-computed candidate embeddings.

To comprehensively evaluate our approach, we conduct extensive experiments under two settings: (1) re-rank evaluation: following previous works \cite{Gu2020SpeakerAwareBF,Su2021DialogueRS,han-etal-2021-fine}, the models are required to rank a given candidate set, and automatic information retrieval metrics are used for evaluation; (2) full-rank evaluation: models are required to search the proper responses from a full candidate pool, and professional human annotators are recruited to measure the quality of the top-1 searched responses. 
Furthermore, in the full-rank evaluation, we also examine the DR-BERT model by adding two nonparallel corpora. To the best of our knowledge, our work is the first effort to select responses from the nonparallel corpus for dialogue response selection.



Experimental results show that our model sets new state of the art on widely used benchmarks under the conventional re-rank setup.
Specifically, compared with the current state-of-the-art cross-encoder model BERT-FP \cite{han-etal-2021-fine}, DR-BERT achieves an absolute improvement in R$_{10}$@1 by 4.14\%, 10.40\%, 13.53\% on Douban, E-commerce, and RRS corpus, respectively. In the full-rank setting, experimental results demonstrate the superiority of DR-BERT over several strong recall-then-rerank baselines as well as the fast ranking baseline.
Moreover, the human evaluations also show that enlarging the candidate pool with the nonparallel corpus that contains large amounts of unpaired sentences leads to better response quality.
In summary, our contributions are:
\begin{itemize}
  \item We apply the dense retrieval model in dialogue response selection and set the new state-of-the-art in both re-rank and full-rank setups. 
  \item More importantly, we show the potential of this approach to select the response from the nonparallel corpus.
  \item Extensive experiments and in-depth analyses are conducted to reveal our approach's merits and inner workings.
  \item We release a high-quality multi-turn dialogue response selection corpus to facilitate future research in this direction.
\end{itemize}

\section{Related Work}
\begin{figure*}[t]     
  \center{\includegraphics[width=\textwidth] {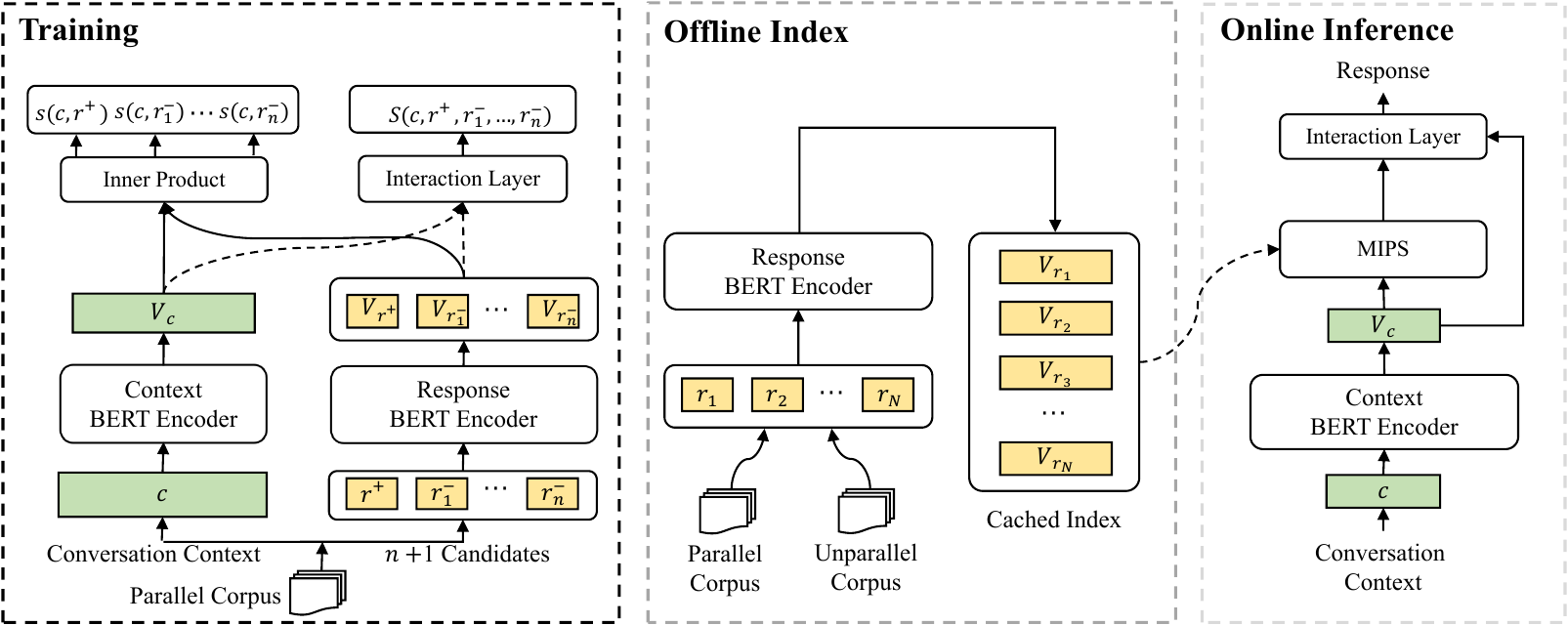}}        
  \caption{
  The overview of our DR-BERT training, offline index, and online inference. $c$ and $r$ represent conversation context and response, respectively. $V_{c}$ and $V_{r}$ denotes their representations in the semantic space. Note that the offline index may come from the nonparallel corpus and thus enables the DR-BERT to select responses from the vast nonparallel resources during online inference.}
  \label{img:overview}
\end{figure*}
\subsection{Dialogue Response Selection Models}

We can divide existing dialogue response selection models into two categories: cross-encoder and dual-encoder models \cite{Humeau2020PolyencodersAA,Tao2021ASO}. They consist of two components: the encoding function $f$ and the aggregation function $\rho$.

\paragraph{Cross-encoder models}
$f$ first encodes the context-response pair $(c, r)$ and collects the rich context-response interaction information. Then, $\rho$ generates the matching score between conversation context and candidate based on the collected interaction information.
For earlier previous works \cite{wu-etal-2017-sequential,Yuan2019MultihopSN}, $f$ is built by RNNs, CNNs, and self-attention networks, and $\rho$ is usually constructed with RNNs. For recent works \cite{Whang2020AnED,Gu2020SpeakerAwareBF,Xu2021LearningAE,Su2021DialogueRS,han-etal-2021-fine}, $f$ is a heavy pre-trained language model (PLM), processing the concatenation of the context and response. $\rho$ is a nonlinear transformation followed by a sigmoid function.


\paragraph{Dual-encoder models}
$f$ first encodes the context and response into their semantic representations separately using neural models. Then, the similarity function $\rho$ (e.g., inner product and cosine similarity) calculates the matching degree between the context and the response.
Previous works \cite{Lowe2015TheUD,Zhou2016MultiviewRS} build $f$ by using CNNs, RNNs, and self-attention networks. Some dense retrieval researches \cite{Karpukhin2020DensePR,Khattab2020ColBERTEA,Santhanam2021ColBERTv2EA,Qu2021RocketQAAO,Ren2021RocketQAv2AJ,Humeau2020PolyencodersAA} recently employed the widely used PLMs (pre-trained language models), i.e., the BERT model, to build the $f$.

\subsection{Dialogue Response Selection Framework}
The most popular and widely used framework for dialogue response selection is the recall-then-rerank pipeline framework \cite{wu-etal-2017-sequential,Fu2020ContexttoSessionMU}, which consists of cascaded recall and re-rank modules. Recall module (e.g., BM25 \cite{Robertson2009ThePR} and docTTTTTquery \cite{Cheriton2019FromDT}) first recalls a coarse-grained candidate set from the whole corpus based on the similarity between a given query and the candidate's context.
Then, the heavy re-rank module (e.g., cross-encoder models) ranks the coarse-grained candidate set, and the dialogue system will return the response with the highest matching degree to the user.
Unlike the conventional pipeline framework that usually contains two models or modules, our DR-BERT could be used to conduct both the recall and re-rank process. Besides, since the interaction layer re-uses the representations provided by the dual-encoder model in DR-BERT, it only introduces very little re-rank computational overhead.

\subsection{Dense Retrieval}
The dense retrieval technique \cite{Karpukhin2020DensePR} has been widely applied in lots of downstream research, such as open-domain question answer and passage retrieval. Many works have proved the effectiveness and efficiency of the dense retrieval models, such as ColBERT \cite{Khattab2020ColBERTEA,Santhanam2021ColBERTv2EA} and RocketQA \cite{Qu2021RocketQAAO,Ren2021RocketQAv2AJ}. 
However, the dense retrieval model has not been well explored in the dialogue response selection to the best of our knowledge.
Whether the dense retrieval models can achieve better performance than the state-of-the-art cross-encoder models remains an open problem in dialogue response selection.

To boost the performance of the dense retrieval models, previous works in dense passage retrieval \cite{Ren2021RocketQAv2AJ,Santhanam2021ColBERTv2EA} propose some effective training strategies to optimize the dense retrieval model: (1) cross-batch negative sampling; (2) denoised hard negative sampling; (3) data augmentation brought from better cross-encoder models; (4) joint training for cross-encoder and dense retrieval models. 
However, we notice that these strategies are limited in the dialogue response selection task. 
This paper employs another three simple yet effective training strategies to optimize the dialogue dense retrieval model, which helps our DR-BERT model significantly outperform the state-of-the-art cross-encoder dialogue response selection models.

\section{Methodology}

In this section, we first introduce the model architecture in Section \ref{sec:dr-bert training}. 
Then, the multi-task training objective is introduced in Section \ref{sec:training_objective}.
Moreover, we present three training strategies to optimize our proposed model effectively in Section \ref{sec:training_strategies}. Finally, the offline index and online inference processes are described in Section \ref{sec:online_inf}.

\subsection{Model Architecture} 
\label{sec:dr-bert training}

Figure \ref{img:overview} shows an overview of the training and inference of the DR-BERT model. 
Our DR-BERT model contains two components: (1) dual-encoder model for fast candidate retrieval; (2) interaction layer for accurate candidate set re-rank.

\subsubsection{Dual-encoder Model}

Our DR-BERT model separately represents the context $c_i$ and response $r_i$ to two vectors $V_{c_i}$ and $ V_{r_i}$ via two decoupled encoders and obtains their matching degree via their inner product as: 
\begin{equation}
  \begin{split}
    & S(c_i, r_i) = V_{c_i}^TV_{r_i}
  \end{split}
  \label{eq:the_first_equ}
\end{equation}
Since the encoders are decoupled, we can therefore pre-compute the representations of all possible responses, including unpaired sentences, and cache them as the index. Then, the response selection can be reduced to Maximum Inner Product Search (MIPS) with the cached index during online inference.  The retrieval can be done efficiently with well-designed data structures and search algorithms \cite{1238663,Norouzi2012FastSI,JDH17} (e.g., inverted index and locality-sensitive hashing).

\subsubsection{Interaction Layer}

\begin{figure}[t]     
  \center{\includegraphics[width=0.35\textwidth] {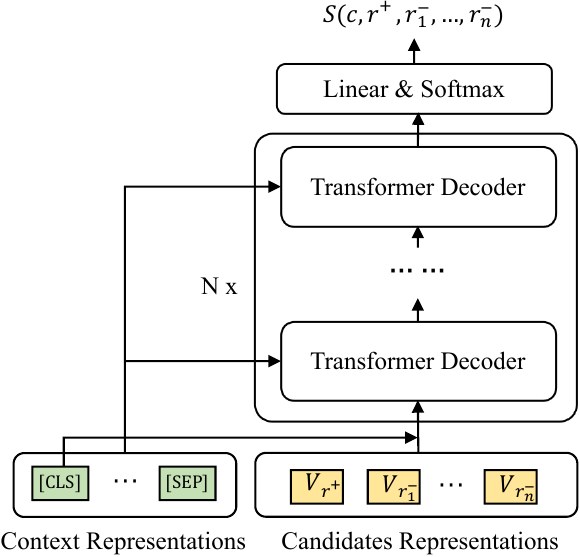}}        
  \caption{The overview of our designed interaction layer. It contains $N$ transformer decoder layers.}
  \label{img:interaction_layer}
  \vspace{-0.5cm}
\end{figure}


The interaction layer aims to rank the whole candidate set directly based on the context-response and response-response interaction information.
This motivation is different from the previous cross-encoder dialogue response selection models \cite{Karpukhin2020DensePR,Khattab2020ColBERTEA,Ren2021RocketQAv2AJ,Su2021DialogueRS,han-etal-2021-fine}, which score each candidate independently, and only consider the context-response interaction information. The interaction among the responses is not considered in these cross-encoder and dense retrieval models, providing richer information for the model to accurately rank the candidate set.

To achieve this goal, we construct the interaction layer with the classic transformer decoder block \cite{Vaswani2017AttentionIA} to model both the context-response and response-response interaction information. The overview of the interaction layer is shown in Figure \ref{img:interaction_layer}.
Suppose that there are hidden states of context $c$ generated by context BERT encoder: $\{V_{c_{j}}\}_{j=0}^m$ ($m$ is the number of the tokens in $c$), and its $n+1$ (1 positive $V_{r^+}$ and $n$ negative $\{V_{r_i^-}\}_{i=1}^n$) candidate representations generated by response BERT encoder.
Specifically, we first augment the candidate representation by adding the [CLS] token embedding of $c$.
Then, the self-attention mechanism in the transformer decoder blocks empowers the DR-BERT model to consider both the context-response and response-response interaction information.
Finally, the linear projection with the Softmax function is used to generate their matching degrees from the final candidate representations $V_{r_i^-}^{'}$ (or $V_{r^+}^{'}$):
\begin{equation}
  \begin{split}
    & S(c, r^+, r_1^-, ..., r_n^-) = {\rm Softmax}(WV_{r_i}^{'} + b)
  \end{split}
  \label{eq:the_first_equ}
\end{equation}
, where $W$ and $b$ is the parameters of the linear projection.

\subsection{Training} 
\label{sec:training_objective}

\subsubsection{Training Objective} 

Our work aims to use one DR-BERT model to conduct the fast recall and accurate re-rank processes. Thus, we design the multi-task training objective to optimize the dual-encoder model and interaction layer in our DR-BERT model. Suppose that there is a positive instance $(c,r^+)$ in the corpus, we select $n$ negative responses $r_i^-$.

To optimize the dual-encoder model, we optimize the loss function for the dual-encoder model as the negative log-likelihood of the positive response:
\begin{equation}
  \begin{split}
    & L_{\rm dual-encoder}=-\frac{1}{n}\sum_{i=1}^n\log\frac{e^{S(c, r^+)}}{e^{S(c, r^+)} +\sum_{j=1}^n e^{S(c, r_j^-)}}
  \end{split}
  \label{eq:negative_log_likelihood}
\end{equation}
To optimize the interaction layer module, we apply the listwise ranking loss function (cross-entropy loss) \cite{Pang2020SetRankLA}:
\begin{equation}
  \begin{split}
    & L_{\rm interaction}=-y\log S(c, r^+, r_1^-, ..., r_n^-)
 \end{split}
  \label{eq:negative_log_likelihood_2}
\end{equation}
, where $y$ denotes the the $n$-dimension one-hot label.
Finally, we add these two training objectives to get the final objective function:
\begin{equation}
  \begin{split}
    & L=L_{\rm dual-encoder} + L_{\rm interaction}
  \end{split}
  \label{eq:negative_log_likelihood_3}
\end{equation}

\subsubsection{Training Strategies} \label{sec:training_strategies}
To push the limits of dense retrieval for dialogue response selection, we employ three simple yet effective training strategies to train the DR-BERT model, which helps the DR-BERT model achieve the state-of-the-art performance in dialogue response selection task: (1) in-batch contrastive learning; (2) pre-train on nonparallel data; (3) data augmentation.

\paragraph{In-batch Contrastive Learning}
\label{sec:inbatch}

To effectively train the DR-BERT model, we employ the in-batch negative sampling (CL) \cite{Humeau2020PolyencodersAA,Karpukhin2020DensePR,DBLP:journals/corr/abs-2111-04198} to optimize the shared semantic space.
Specifically, in a training batch $B$ of size $n$, any $(c_i, r_j)$ pair is a positive sample when $i = j$, and negative otherwise. In this way, we re-use the computation and effectively train on $n^2$ $(c_i, r_j)$ pairs in each batch. In other words, the model will be trained on $n$ times as many negative samples as before. Given a batch of context-response pairs $\{(c_i,r_i)\}_{i=1}^n$, the Eq. (\ref{eq:negative_log_likelihood}) can be rewritten to:
\begin{equation}
  \begin{split}
    & L=-\frac{1}{n}\sum_{i=1}^n\log\frac{e^{S(c_i,r_i)}}{\sum_{j=1}^n e^{S(c_i,r_j)}}
  \end{split}
\end{equation}
Although the cross-batch negative sampling has been proved effective in previous works in passage retrieval \cite{Qu2021RocketQAAO,Santhanam2021ColBERTv2EA}, we don't notice the performance gain from cross-batch negative sampling in the dialogue response selection.

\paragraph{Pretrain on Nonparallel Data}

Our proposed DR-BERT model can potentially select responses from the nonparallel corpus by the dual-encoder model, including unpaired sentences. 
However, since these sentences are not used in training, their representation in the cached index may not be accurate.


To address this problem, the \textbf{D}omain \textbf{A}daptive \textbf{P}retraining is proposed to warm up the DR-BERT model on the \textbf{N}onparallel corpus, namely NDAP. Specifically, we perform the RoBERTa-style pretraining \cite{Liu2019RoBERTaAR} on the nonparallel corpus before fine-tuning, which optimizes the masked language model (MLM) objective with dynamic masking.


Furthermore, previous domain adaptive pretraining methods \cite{Whang2020AnED,han-etal-2021-fine} need large amounts of parallel context-response pairs to warm up the BERT model, for example, the BERT-FP domain adaption pretraining \cite{han-etal-2021-fine}. On the contrary, our NDAP methods could liberate from the limits of the parallel corpus and obtain huge improvements from the easy-to-collect nonparallel corpus, even the non-conversational corpus. In this paper, we don't leverage the nonparallel corpus to boost the NDAP's performance for fair comparisons.

\paragraph{Data Augmentation}

To collect the high-quality corpus for DR-BERT training, we design the data augmentation technique for multi-turn dialogue response selection, which contains two parts.

(1) positive instances expansion: since the dialogue response selection corpus usually contains the multi-turn conversation sessions, it is very straightforward to expand the size of the positive instances for more robust DR-BERT fine-tuning. Specifically, given one multi-turn conversation context $[c_{i,1},c_{i,2},...,c_{i,m_i}]$ of turn length $m_i$, we could split it into $k$ training samples $\{([c_{i,1},...,c_{i,m_i-j}],$ $c_{i,m_i-j+1})\}_{j=1}^k$. As $k$ increases, more short and fine-grained training samples will be created.

(2) hard negative selection: prior works in passage retrieval and dialogue response selection \cite{Ren2021RocketQAv2AJ,Su2021DialogueRS,han-etal-2021-fine} prove the importance of the hard negative instances during training. To obtain the hard negative samples, they usually conduct the recall-then-rerank pipeline to select the proper hard negative candidates for each context. However, their methods are very time-consuming and computation-intensive. Besides, recall-then-rerank pipelines usually introduce the false negative samples due to the common one-to-many relationship in the dialogue sessions \cite{Bao2020PLATOPD}. For the dialogue response selection, we notice that these false negative samples decrease the performance badly.
In this paper, we choose the weak BM25 module to construct the hard negative instances, and two kinds of selection strategies are used:
(a) context-context selection: the whole multi-turn conversation context is used to search the similar conversation context, and its corresponding response is chosen as the hard negative samples;
(b) context-response selection: each utterance in the multi-turn conversation context will be used to directly search the similar utterances as the hard negative samples.
The hard negative samples searched by our methods are weaker than the recall-then-rerank data augmentation technique, which is beneficial for the robust training.

\subsection{Offline Index and Online Inference}
\label{sec:online_inf}
\paragraph{Offline Index}
After training the DR-BERT model, we can conduct efficient online inference easily. First of all, we need to conduct the following steps to build the offline index. As shown in the middle part of Figure \ref{img:overview}, given the unpaired responses $\{r_i\}_{i=1}^{N}$ collected from the parallel and nonparallel resources, where $N$ is the number of the responses, the response encoder of DR-BERT converts them into corresponding semantic embeddings $\{V_{r_i}\}_{i=1}^{N}$. Then, the cached index can be built with the original sentences and their corresponding embeddings $\{(V_{r_i},r_i)\}_{i=1}^{N}$. 

\paragraph{Online Inference}
The online inference process is shown in the right part of Figure \ref{img:overview}. When we receive a dialogue history $C$, the context encoder of DR-BERT first encodes it into the context semantic embedding $V_{c}$. Then, the maximum inner product search (MIPS) is used to find candidates $\{r_i\}_{i=0}^N$ that have high matching degrees with $C$.
Finally, the interaction layer rank the whole candidate set given the context $C$, and generates their final matching degrees $S(C, \{r_i\}_{i=0}^n)^{\rm inter.}$. The response with the highest matching degree will be returned to the user.

\subsection{Implementation Details}

Our codes are based on PyTorch deep learning framework \cite{Paszke2019PyTorchAI} and the Huggingface package \cite{wolf-etal-2020-transformers}. 
The experiments, including domain adaptive pretraining and fine-tuning, run on 8 Tesla V100 GPUs with 32 Gb memory.
For DR-BERT training, following the previous works, we use the bert-base-chinese checkpoint for Chinese datasets and the bert-base-uncased checkpoint for the English dataset. 
These checkpoints are warmed up by the BERT-FP \cite{han-etal-2021-fine} pretraining method and our proposed nonparallel domain adaptive pretraining strategy in this paper.
The transformer layer in our interaction layer is 2 for all the datasets. The number of the head in the transformer layer is 12, and the hidden size is 768.
The fine-grained degree $k$ in our proposed fine-grained data augmentation strategy are 10 for all of the datasets. 
The epoch is 5 for Douban, Ubuntu, and RRS datasets 10 for E-commerce corpus.
The batch size is 64 for all the datasets.
The maximum number of the tokens in the dialogue context and response is (256, 64) for E-commerce, Douban, Ubuntu datasets, and RRS corpus.
The DR-BERT model is optimized by AdamW optimizer \cite{Kingma2015AdamAM} with a learning rate of 5e-5, and the linear learning ratio scheduler is used.
We choose the FAISS package \cite{JDH17} to build and search the cached index for online inference. 
We choose the very famous and widely used Elasticsearch toolkit (Lucene-BM25 system) for BM25 recall.



\section{Experiments}
We test our proposed DR-BERT model under two evaluation settings: re-rank and full-rank evaluation. We follow the conventional evaluation protocol \cite{Gu2020SpeakerAwareBF,Su2021DialogueRS,han-etal-2021-fine} for the re-rank evaluation, which tests whether the baseline models can accurately rank a small candidate set based on their corresponding dialogue context. For the full-rank evaluation, the models are required to select proper responses from the whole corpus, and the quality of the top-1 searched responses is measured by human evaluation. Nevertheless, the conventional benchmark datasets are either domain-specific or contain many ambiguous cases, which brings difficulty to our human annotators in the full-rank evaluation. Therefore, to facilitate future research in this direction, we build a new high-quality multi-turn dialogue response selection corpus, named \textbf{R}estoration200k for \textbf{R}esponse \textbf{S}election (RRS). We then conduct both re-rank and full-rank evaluations on this newly-introduced dataset.

\begin{table*}[tb]
\small
\renewcommand{\arraystretch}{1.2}
	\setlength{\tabcolsep}{3.2pt}
	\scalebox{0.83}{
      \begin{tabular}{ccccccccccccccccccc}
      \hlinewd{0.75pt}
      \multirow{2}{*}{\textbf{Models}}&\multicolumn{6}{c}{\textbf{Douban}}&\multicolumn{3}{c}{\textbf{Ubuntu}}&\multicolumn{6}{c}{\textbf{RRS}}&\multicolumn{3}{c}{\textbf{E-commerce}}\\ 
      \cmidrule(lr){2-7}
      \cmidrule(lr){8-10}
      \cmidrule(lr){11-16}
      \cmidrule(lr){17-19}
      &MAP&MRR&P@1&R$_{10}$@1&R$_{10}$@2&R$_{10}$@5&\rm R$_{10}$@1&\rm R$_{10}$@2&\rm R$_{10}$@5&MAP&MRR&P@1&R$_{10}$@1&R$_{10}$@2&R$_{10}$@5&R$_{10}$@1&R$_{10}$@2&R$_{10}$@5\\ 
      \hlinewd{0.75pt}
      DualLSTM&-&-&-&-&-&-&0.638&0.784&0.949&-&-&-&-&-&-&0.365&0.536&0.828\\ 
      Multi-View&-&-&-&-&-&-&0.662&0.801&0.951&-&-&-&-&-&-&0.421&0.601&0.861\\ 
      SMN&0.529&0.569&0.397&0.233&0.396&0.724&0.726&0.847&0.961&0.487&0.501&0.309&0.281&0.442&0.723&0.453&0.654&0.886\\ 
      DUA&0.551&0.599&0.421&0.243&0.421&0.780&0.752&0.868&0.962&-&-&-&-&-&-&0.501&0.700&0.921\\  
      DAM&0.550&0.601&0.427&0.254&0.410&0.757&0.767&0.874&0.969&0.511&0.534&0.347&0.308&0.457&0.751&0.526&0.727&0.933\\ 
      IoI&0.573&0.621&0.444&0.269&0.451&0.786&0.796&0.894&0.974&-&-&-&-&-&-&0.563&0.768&0.950\\ 
      ESIM&-&-&-&-&-&-&0.796&0.874&0.975&-&-&-&-&-&-&0.570&0.767&0.948\\ 
      MSN&0.587&0.632&0.470&0.295&0.452&0.788&0.800&0.899&0.978&0.550&0.563&0.383&0.343&0.498&0.798&0.606&0.770&0.937\\  
      IMN&0.570&0.615&0.433&0.262&0.452&0.789&0.794&0.889&0.974&-&-&-&-&-&-&0.621&0.797&0.964\\
      \hline
      BERT&0.591&0.633&0.454&0.280&0.470&0.828&0.817&0.904&0.977&0.625&0.639&0.453&0.404&0.606&0.875&0.610&0.814&0.973\\
      SA-BERT&0.619&0.659&0.496&0.313&0.481&0.847&0.855&0.928&0.983&0.660&0.670&0.488&0.444&0.653&0.922&0.704&0.879&0.985 \\
      Poly-encoder&0.608&0.650&0.475&0.299&0.494&0.822&0.882&0.949&0.990&0.715&0.729&0.578&0.518&0.708&0.925&0.924&0.963&0.992\\
      ColBERT&0.608&0.649&0.471&0.296&0.492&0.838&0.830&0.910&0.978&0.692&0.706&0.555&0.501&0.656&0.915&0.871&0.938&0.990 \\
     MDFN&0.624&0.663&0.498&0.325&0.511&0.855&0.866&0.932&0.984&-&-&-&-&-&-&0.639&0.829&0.971\\
     UMS$_{\rm BERT+}$&0.625&0.664&0.499&0.318&0.482&0.858&0.876&0.942&0.988&-&-&-&-&-&-&0.762&0.905&0.986\\ 
     BERT-SL&-&-&-&-&-&-&0.884&0.946&0.990&-&-&-&-&-&-&0.776&0.919&0.991\\ 
     SA-BERT+HCL&0.639&0.681&0.514&0.330&0.531&0.858&0.867&0.940&0.992&0.671&0.683&0.503&0.454&0.659&0.917&0.721&0.896&0.993\\
     BERT-FP$^{\dagger}$&0.644&0.680&0.512&0.324&0.542&0.870&\textbf{0.911}&\textbf{0.962}&\textbf{0.994}&0.709&0.724&0.565&0.505&0.705&0.932&0.870&0.956&0.993\\ 
     
     \hline
     DR-BERT &\textbf{0.659}&\textbf{0.695}&\textbf{0.520}&\textbf{0.338}&\textbf{0.572}&\textbf{0.880}&0.910&\textbf{0.962}&\textbf{0.993}&\textbf{0.758}&\textbf{0.771}&\textbf{0.648}&\textbf{0.584}&\textbf{0.744}&0.928&\textbf{0.971}&\textbf{0.987}&\textbf{0.997} \\
     w/o. IL&0.648&0.685&0.516&0.331&0.550&0.868&\textbf{0.913}&\textbf{0.961}&\textbf{0.993}&0.733&0.746&0.606&0.542&0.727&\textbf{0.933}&0.960&0.984&0.996 \\
     w/o. NDAP &0.633&0.672&0.498&0.319&0.529&0.851&0.905&0.957&0.992&0.739&0.753&0.620&0.557&0.721&0.919&0.949&0.984&\textbf{0.997}\\ 
     w/o. DA&0.613&0.655&0.496&0.311&0.496&0.834&0.889&0.950&0.991&0.712&0.726&0.573&0.512&0.705&0.917&0.925&0.969&0.995\\ 
     w/o. CL&0.616&0.655&0.487&0.309&0.501&0.819&0.888&0.943&0.988&0.678&0.690&0.540&0.484&0.655&0.888&0.891&0.955&0.991\\
      \hlinewd{0.75pt}
      \end{tabular}}
    \caption{The re-rank experimental results and ablation study on Douban corpus, Ubuntu corpus, E-commerce corpus, and our released RRS corpus. IL, NDAP, DA, and CL represent interaction layer, nonparallel domain adaptive pretraining, data augmentation technique, and in-batch contrastive learning. IL denotes the interaction layer. ${\dagger}$ denotes the state-of-the-art cross-encoder dialogue response selection model.
    }
    \label{tab:main_result}
    \vspace{-0.6cm}
\end{table*}
\subsection{The RRS Dataset} The RRS corpus is built from a Chinese high-quality open-domain dialogue dataset, Restoration200K \cite{pan-etal-2019-improving}. 
Human annotators examine all dialogue sessions in Restoration200K to ensure readability.
Following the previous works \cite{wu-etal-2017-sequential}, we further process this dataset for response selection task: (1) \textit{Train set}: for each context-response pair in the train set, we randomly sample response in the train set as its negative sample; (2) \textit{Validation set}: given one context-response pair, we collect 9 extra hard negative samples by using BM25 recall module; 
(3) \textit{Test set}: we first sample 1,200 context-response pairs. Then, we search 15 hard negative samples for each multi-turn context by using the BM25 recall module. 
To ensure our test set's quality, we hire three professional annotators to re-label these 16 candidates (1 ground-truth and 15 hard negative samples) for each context.
So the ``false negative'' samples in 15 hard negative samples can be effectively detected.
Specifically, annotators are required to classify each candidate into three categories: positive, negative, and ambiguous.
If one annotator classifies the candidate into the ambiguous one, we remove it from the candidate set.  
Then, each valid candidate receives three labels from three annotators, and the majority of the labels are taken as the final decision. After discarding the sessions lacking negative samples, we keep 500 valid sessions as the test set, and each session consists of 10 candidates. Note that Fleiss's Kappa \cite{Fleiss1971MeasuringNS} of the labeling is 0.53,  which indicates relatively high agreement among annotators.

\subsection{Re-rank Evaluation}
\subsubsection{Datasets} \label{exp:re-rank-dataset}
We test baselines on the four response selection datasets:
\begin{itemize}[wide=0\parindent,noitemsep,topsep=0em]
  \item \textbf{Ubuntu Corpus} Ubuntu IRC Corpus V1 \cite{Lowe2015TheUD} 
  is a publicly available English domain-specific dialogue dataset.
  \item \textbf{Douban Corpus} Douban Corpus \cite{wu-etal-2017-sequential}
  is a widely used Chinese multi-turn open-domain dialogue dataset, collected from the Douban group, a popular social networking service.
  \item \textbf{E-commerce Corpus} E-commerce Corpus \cite{zhang-etal-2018-modeling}
  is a Chinese domain-specific multi-turn dialogue dataset collected from the largest e-commerce platform, Taobao. 
  It contains real-world conversations between customers and service.
  \item \textbf{RRS Corpus} 
  The datasets mentioned above are either domain-specific or contain many ambiguous cases, which makes our human annotations difficult. On the contrary, all of the dialogue sessions in our released RRS dataset are annotated by humans, which is more high-quality than previous datasets.
\end{itemize}

\begin{table}[t]
\tiny
\renewcommand{\arraystretch}{1.2}
  \begin{center}
        \resizebox{0.45\textwidth}{!}{
          \begin{tabular}{c|ccc|ccc}
          \hlinewd{0.7pt}
          \multirow{2}{*}{\textbf{Dataset}} 
          & \multicolumn{3}{c|}{\textbf{Ubuntu}}          
          & \multicolumn{3}{c}{\textbf{Douban}}           \\ 
          \cline{2-7}
          &Train&Val&Test&Train&Val&Test\\ \hline
          size      &1M&500K&500K&1M&50K&10K\\ 
          pos:neg   &1:1&1:9&1:9&1:1&1:1&1.2:8.8 \\ 
          avg turns &10.13&10.11&10.11&6.69&6.75&6.45 \\ 
          \hlinewd{0.75pt}
          \end{tabular}
        }
    \qquad
        \resizebox{0.45\textwidth}{!}{
          \begin{tabular}{c|ccc|ccc}
          \hlinewd{0.7pt}
          \multirow{2}{*}{\textbf{Dataset}} 
          & \multicolumn{3}{c|}{\textbf{E-commerce}}      
          & \multicolumn{3}{c}{\textbf{RRS}}  \\ 
          \cline{2-7}
          &Train&Val&Test&Train&Val&Test\\ \hline
          size      &1M&10K\ \ &\ \ 10K\ \ \ &0.4M&50K&5K\\ 
          pos:neg   &1:1&1:1&1:9&1:1&1:9&1.2:8.8 \\ 
          avg turns &5.51\ &5.48\ &5.64\ &5&5&5 \\ 
          \hlinewd{0.75pt}
          \end{tabular}
        }
    \caption{The statistics of four multi-turn response selection datasets that are used in this paper.}
    \label{tab:dataset}
  \end{center}
  \vspace{-1.0cm}
\end{table}

For all of the datasets, the nonparallel training corpus that is used to conduct the nonparallel domain adaption pretraining (NDAP) consists of the single sentences collected from their train set. 
The statistics of these four datasets are shown in Table \ref{tab:dataset}.

\subsubsection{Evaluation Metrics}
Following previous works \cite{Xu2016IncorporatingLK,wu-etal-2017-sequential,zhang-etal-2018-modeling,pan-etal-2019-improving,han-etal-2021-fine}, we use a set of automatic evaluation metrics to test baselines: (1) recall at position k in 10 candidates (R$_{10}$@k); (2) mean average precision (MAP); (3) mean reciprocal rank (MRR); (4) precision at one (P@1).

\subsubsection{Baselines}
According to whether to use the pretrained language models (PLMs) \cite{devlin-etal-2019-bert,Liu2019RoBERTaAR}, we divide the previous works into two categories: Non-PLM-based models and PLM-based models.
\begin{itemize}[wide=0\parindent,noitemsep,topsep=0em]
\item \textbf{Non-PLM-based Matching Models}
Before PLMs became dominant in NLP research, previous works constructed matching models using CNNs, RNNs, and stacked self-attention networks.
This kind of model includes, but is not limited to, DualLSTM \cite{Lowe2015TheUD}, Multi-View \cite{Zhou2016MultiviewRS}, SMN \cite{wu-etal-2017-sequential}, DUA \cite{zhang-etal-2018-modeling}, DAM \cite{Zhou2018MultiTurnRS}, IoI \cite{Tao2019OneTO}, ESIM \cite{Chen2019SequentialMM}, MSN \cite{Yuan2019MultihopSN}, MRFN \cite{Tao2019MultiRepresentationFN}, IMN \cite{Gu2020UtterancetoUtteranceIM}.
\item \textbf{PLM-based Matching Models} Recently, with the wide application of PLMs in many downstream NLP tasks, more and more PLM-based matching models, such as cross-encoder and dense retrieval models, are proposed.
With the help of the powerful natural language understanding capability in PLMs, recent researches refresh the state-of-the-art performance repeatedly. This kind of model includes, but is not limited to, BERT \cite{Whang2020AnED}, SA-BERT \cite{Gu2020SpeakerAwareBF},
MDFN \cite{liu2021filling}, 
UMS$_{\rm BERT+}$ \cite{Whang2021DoRS}, BERT-SL \cite{Xu2021LearningAE}, SA-BERT+HCL \cite{Su2021DialogueRS}, BERT-FP \cite{han-etal-2021-fine}.  Note that BERT-FP is the current state-of-the-art cross-encoder dialogue response selection model. We also test the popular and powerful dense retrieval baselines Poly-encoder \cite{Humeau2020PolyencodersAA} and ColBERT \cite{Santhanam2021ColBERTv2EA}, which is never considered in previous dialogue response selection works. Since the false negative problem in dialogue response selection is more severe than in passage retrieval, we only supply the hard negative samples for ColBERT retrieved by the BM25 recall module. Besides, the in-batch negative sampling described in ColBERTv2 \cite{Santhanam2021ColBERTv2EA} model is also used for ColBERT model.
\end{itemize}

\subsubsection{Experimental Results}



The re-rank experimental results are shown in Table \ref{tab:main_result}. We can see that the proposed DR-BERT model achieves comparable, if not better, performance than state-of-the-art cross-encoder models (e.g., BERT-FP). In fact, DR-BERT establish new state-of-the-art on most datasets and the p-value of the significance test is much lower than 0.01. This indicates the improvement of DR-BERT is substantial.
To reveal that DR-BERT benefits a lot from our training strategies, we conduct a set of ablation study to analyze the contribution individual training strategies. The results of the ablation study are shown in the last three rows of Table \ref{tab:main_result}.
It can be observed that the performance decreases if any training strategy is removed. This observation suggests that the proposed training strategies are complementary to each other. Besides, it can be seen that removing the interaction layer leads to performance degradation, which proves its effectiveness. It is worth noting that even if the interaction layer (refer to the w/o. IL in the last block of Table \ref{tab:main_result}) is removed, our DR-BERT model still outperforms the previous state-of-the-art cross-encoder models, which reveals the huge potential of the dense retrieval model in dialogue response selection task.

We also conduct comparison experiments to deeply analyze the difference between our training strategies with previous works.

The comparisons of domain adaption technique are shown in Table \ref{tab:ablation_ndap_cl} (a), from which we can make the following conclusions:
(1) it can be found that both BERT-FP and our NDAP domain adaption techniques bring huge improvement for the DR-BERT model;
(2) DR-BERT$_{\rm NDAP}$ performance is better than DR-BERT$_{\rm BERT-FP}$, which demonstrates that our NDAP domain adaption pretraining is more suitable than previous state-of-the-art domain adaption pretraining in dialogue response selection.

The comparisons of training methods are shown in Table \ref{tab:ablation_ndap_cl} (b). It could be found that the performance of cross-batch negative sampling and in-batch negative sampling is very close. The phenomenon demonstrates that this powerful training strategy in passage retrieval \cite{Qu2021RocketQAAO} is limited in dialogue response selection.

The analysis of the data augmentation techniques are shown in Table \ref{tab:ablation_ndap_cl} (c), from which we could make the following conclusions:
(1) the proposed data augmentation technique significantly improves the DR-BERT's performance;
(2) in our data augmentation technique, the hard negative selection contributes more to the DR-BERT's improvement than the positive instances expansion.



\begin{table}[h]
\small
\begin{center}
\renewcommand{\arraystretch}{1.2}
	\setlength{\tabcolsep}{3.2pt}
	\subtable[Comparison of Domain Adaption Pretraining: None, BERT-FP, NDAP denote that no domain adaption, BERT-FP domain adaption \cite{han-etal-2021-fine}, and our proposed NDAP domain adaption are used to warm up the BERT models, respectively.]{
    	\scalebox{0.95}{
            \begin{tabular}{ccccccc}
            \hlinewd{0.75pt}
            \textbf{Models}          & \textbf{MAP} & \textbf{MRR} & \textbf{P@1} & \textbf{R$_{10}$@1} & \textbf{R$_{10}$@2} & \textbf{R$_{10}$@5} \\ \hlinewd{0.75pt}
            \textbf{DR-BERT$_{\rm None}$}   & 0.739             &  0.753            &  0.620            &       0.557         &  0.721              &   0.919             \\ 
            \textbf{DR-BERT$_{\rm BERT-FP}$} &  \textbf{0.761}            &  0.747            &  0.628            &       0.564         &   \textbf{0.746}             &  0.922              \\
            \textbf{DR-BERT$_{\rm NDAP}$} & 0.758             &     \textbf{0.771}         &      \textbf{0.648}        &  \textbf{0.584}              &    0.744            &     \textbf{0.928}           \\ \hlinewd{0.75pt}
            \end{tabular}
        }
    }
    \qquad
    \subtable[Comparison of Training Method: in-batch and cross-batch denote that in-batch and cross-batch \cite{Qu2021RocketQAAO} negative sampling training methods.
    ]{
        \scalebox{0.95}{
        \begin{tabular}{ccccccc}
        \hlinewd{0.75pt}
        \textbf{Models}          & \textbf{MAP} & \textbf{MRR} & \textbf{P@1} & \textbf{R$_{10}$@1} & \textbf{R$_{10}$@2} & \textbf{R$_{10}$@5} \\ \hlinewd{0.75pt}
        \textbf{dual-encoder$_{\rm in-batch}$}   &\textbf{0.749}& \textbf{0.766}& \textbf{0.636}&0.567             & 0.739             &     \textbf{0.943}                      \\ 
        \textbf{dual-encoder$_{\rm cross-batch}$} &  \textbf{0.749}            &  0.764            &   0.634           &          \textbf{0.570}      & \textbf{0.744}               & 0.937               \\ \hlinewd{0.75pt}
        \end{tabular}
        }
    }
    \qquad
    \subtable[Analysis of Data Augmentation Technique: w/o. DA, w/o. positive, and w/o. negative denote that no data augmentation, only hard negative selection, and only positive instances expansion are used to boost DR-BERT performance, respectively.]{
        \scalebox{0.95}{
        \begin{tabular}{ccccccc}
        \hlinewd{0.75pt}
        \textbf{Models}          & \textbf{MAP} & \textbf{MRR} & \textbf{P@1} & \textbf{R$_{10}$@1} & \textbf{R$_{10}$@2} & \textbf{R$_{10}$@5} \\ \hlinewd{0.75pt}
        \textbf{DR-BERT}   &\textbf{0.758}&\textbf{0.771}&\textbf{0.648}&\textbf{0.584}&0.744&0.928                      \\ 
        \textbf{DR-BERT$_{\rm w/o.\ positive}$}   &0.746&0.760&0.628&0.567&\textbf{0.745}&0.926                      \\ 
        \textbf{DR-BERT$_{\rm w/o.\ negative}$} &0.746&0.761&0.636&0.568&0.728&\textbf{0.938}              \\
        \textbf{DR-BERT$_{\rm w/o.\ DA}$} &0.712&0.726&0.573&0.512&0.705&0.917              \\\hlinewd{0.75pt}
        \end{tabular}
        }
    }
    \end{center}
    \caption{Analysis on difference between our training strategies with previous works. Experiments are conducted on RRS corpus.} 
    \label{tab:ablation_ndap_cl}
    \vspace{-0.6cm}
\end{table}

\subsection{Full-rank Evaluation} \label{sec:full-rank-experiment}
In full-rank evaluation, the models are required to select proper responses from the whole corpus. 
Note that the full-rank evaluation is conducted on our newly introduced RRS corpus.
We design two different experimental settings for the full-rank evaluation: base and nonparallel settings.
\begin{itemize}[wide=0\parindent,noitemsep,topsep=0em]
    \item \textbf{Base setting} The candidate pool only contains the responses in the train set.
    \item \textbf{Nonparallel setting} A nonparallel corpora is added to enlarge the candidate pool.
    
\end{itemize}

\subsubsection{Evaluation Metrics}
Since the one-to-many relationship is very common in dialogue \cite{Bao2020PLATOPD}, the ground-truth response in dataset may be neither the only proper response nor the best response for the given conversation context. 
Thus, the automatic evaluation metrics are not applicable in the full-rank dialogue response selection.
Referring to the human evaluation metric in dialogue generation \cite{Li2016DeepRL,Adiwardana2020TowardsAH,Roller2021RecipesFB}, we apply human annotation to evaluate the performance fairly and accurately.
Specifically, we recruit 8 professional annotators to evaluate the quality of the top-1 response selected by different models. They are asked to label all dialogue sessions in the test set of RRS. Based on the correlation between the conversation context and responses, they are required to rate each candidate on a five-point scale:
\begin{itemize}[wide=0\parindent,noitemsep,topsep=0em]
\item \textbf{Score 1} It covers the following situations: (1) candidate is completely uncorrelated with the context; (2) the context is ambiguous.
\item \textbf{Score 2} It covers the following situations: (1) candidate has logical contradiction to the context; (2) candidate has a weak correlation with the context; (3) candidate's quality is better than score 1 but worse than score 3.
\item \textbf{Score 3} It covers the following situations: (1) candidate has an ordinary correlation with the conversation context; (2) candidate is very similar to the context; (3) candidate is a safe and general response without any additional information.
\item \textbf{Score 4} The candidate's quality is better than score 3 but worse than score 5.
\item \textbf{Score 5} The candidate is informative, accurate, and has a very high correlation with the context.
\end{itemize}

Note that the Fleiss's Kappa \cite{Fleiss1971MeasuringNS} of this five-point scale labeling is close to 0.4, which indicated relatively high agreement among annotators.

\subsubsection{Baselines}
We choose the following baselines for full-rank evaluation. Note that the candidate set size is 200 for all of the pipeline baselines.
\begin{itemize}[wide=0\parindent,noitemsep,topsep=0em]
\begin{sloppypar}
    \item docTTTTTquery \cite{Cheriton2019FromDT}: the doc2query coarse-grained recall module is used to search the proper responses, which significantly outperforms the conventional BM25 system \cite{Qu2021RocketQAAO,Ren2021RocketQAv2AJ};
    \item docTTTTTquery+ BERT-FP \cite{han-etal-2021-fine}: it jointly employs the 
    docTTTTTquery as the coarse-grained recall module and the BERT-FP as re-rank module to select the response from the whole dialogue corpus;
    \item docTTTTTquery+ poly-encoder \cite{Humeau2020PolyencodersAA}: it substitutes the BERT-FP in the former method to the poly-encoder model, which runs much faster;
    \item docTTTTTquery+ DR-BERT: it uses the docTTTTTquery as the recall module, and the DR-BERT as re-rank module;
    \item ColBERT \cite{Khattab2020ColBERTEA}: ColBERT introduces a late interaction architecture that employs the scalable MaxSim-based interaction on token representations between the context and candidates. Following the ColBERT, we employ it in both recall and re-rank phrases;
    \item DR-BERT: it employs DR-BERT in both recall and re-rank phrase.
\end{sloppypar}
\end{itemize}







\subsubsection{Experimental Results}
\paragraph{Base Setting}
In the base setting, all methods select responses from the RRS train set. Its results are shown in Table \ref{tab:full_rank}, from which we can draw the following conclusions:



(1) The average human scores of docTTTTTquery+DR-BERT significantly outperforms other docTTTTTquery-based pipeline baselines, which means that the DR-BERT model could rank the candidates more accurately than other strong re-rank baselines.
The gap between BERT-FP and poly-encoder is big, but the performance of docTTTTTquery+poly-encoder and docTTTTTquery+DR-BERT are similar. Given that BERT-FP is much slower than poly-encoder and DR-BERT (refer to Section \ref{sec:Inference Speed} for detailed comparison), this result is beyond our expectation. After our careful check, we suggest that this is because some hard negative candidates could easily confuse the cross-encoder models, and obtain higher scores than ground-truths. This observation also proves the vulnerable performance of cross-encoders.

(2) The performance of ColBERT is comparable with the pipeline baselines that contain the docTTTTTquery recall module. Considering that ColBERT's space footprint for the cached index is much bigger than the docTTTTTquery coarse-grained recall module and our proposed DR-BERT model, it is not an excellent good choice to employ the ColBERT in dialogue response selection.

(3) DR-BERT model achieves the best average human scores in the full-rank evaluation and significantly outperforms these strong pipeline baselines. Compared with docTTTTTquery+DR-BERT, after substituting the recall module to the dual-encoder model in DR-BERT, the performance of DR-BERT is significantly improved. It demonstrates that the DR-BERT recalls more proper candidates for further re-ranking.
 


\begin{table}[tb]
\small
\renewcommand{\arraystretch}{1.2}
\setlength{\tabcolsep}{3.2pt}
  \begin{center}
    \scalebox{0.9}{
      \begin{tabular}{cc}
      \hlinewd{0.75pt}
      \textbf{Baselines}      & \textbf{Avg. Human Scores (1-5)} \\ \hlinewd{0.75pt}
      docTTTTTquery           &  2.12                          \\ 
      docTTTTTquery+BERT-FP   &  2.80                      \\ 
      docTTTTTquery+poly-encoder  &  2.92                         \\ 
      docTTTTTquery+DR-BERT   &   2.96                        \\ 
      ColBERT                 &   2.92                        \\ 
      DR-BERT                 &   \textbf{3.15}                          \\ \hlinewd{0.75pt} 
      DR-BERT+in-dataset      &   3.20       (+1.56\%)           \\ 
      DR-BERT+out-dataset     &   \textbf{3.24}  (+2.78\%) 
      \\\hlinewd{0.75pt}
      \end{tabular}
    }
    \caption{Full-rank experimental results on our released high-quality RRS test set. 
    }
    \label{tab:full_rank}
  \end{center}
  \vspace{-1.0cm}
\end{table}

\paragraph{Nonparallel Setting}
In the nonparallel setting,
two nonparallel corpora are added to the candidate pool separately: (1) in-dataset nonparallel corpus contains 0.67 million single sentences collected from the multi-turn conversation context in the RRS training set; (2) out-dataset nonparallel corpus contains over 3.75 million unpaired single sentences crawled from the Douban group, a famous Chinese social website.

The experimental results of the nonparallel setting are shown in the last two rows in Table \ref{tab:full_rank}.
It can be observed that both of the nonparallel corpora notably improve the response quality. DR-BERT+in-dataset and DR-BERT+out-dataset achieve relative improvement of 1.56\% and 2.78\% in average human scores, respectively. 
Besides, we observe that 74.8\% and 84.2\% selected responses of DR-BERT+in-dataset and DR-BERT+out-dataset are from the nonparallel corpora. This result proves that the improvements are brought from them. In addition, DR-BERT+out-dataset is better than DR-BERT+in-dataset for the average human scores, suggesting that increasing the size of the nonparallel corpus is beneficial for the performance.

The results of our investigated nonparallel settings suggest that the DR-BERT model can liberate from the requirement of the parallel corpus in real-world applications and achieve better performance with the easy-to-collect nonparallel resources (e.g., the non-conversational corpus).

\section{Hyper-parameter Analysis}

In this part, we study the two main hyper-parameters in our work: the number of the transformer layer in the interaction layer and fine-grained degree $k$ in the data augmentation technique (positive instances expansion). Note that this test are conducted on the RRS corpus.

\begin{figure}[tb]
  \center{\includegraphics[width=0.48\textwidth, height=6.8cm] {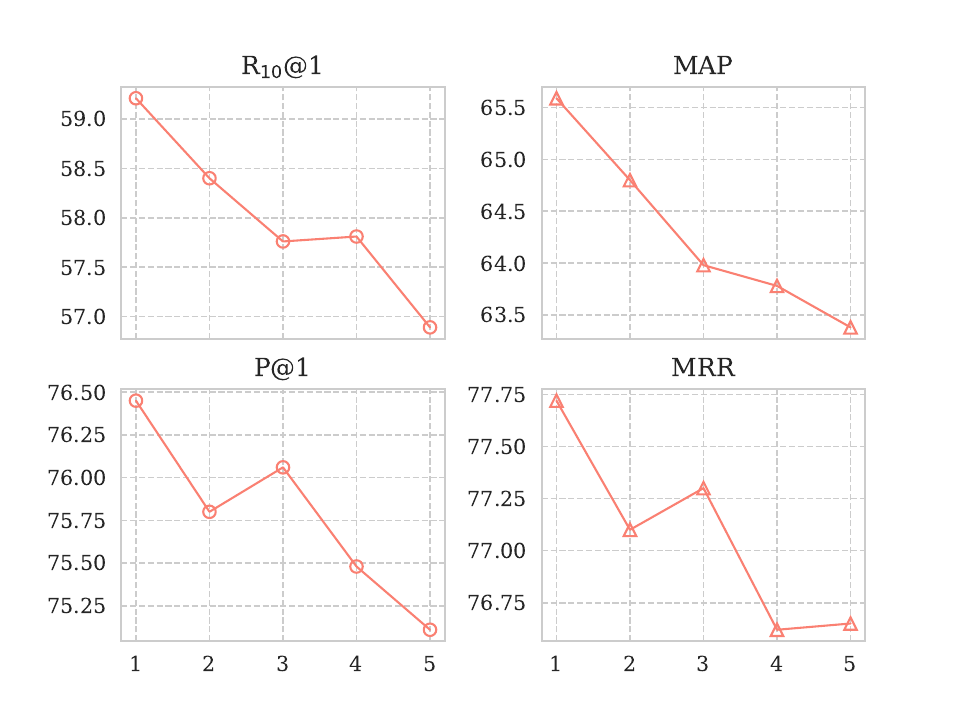}}
  \caption{The effect of the number of the transformer layer. Except for the number of the transformer layer, other hyper-parameters keep the same.
  }
  \label{img:hyper_layer}
  \vspace{-0.5cm}
\end{figure}

\paragraph{Transformer Layer} As shown in Figure \ref{img:hyper_layer}, it can be observed that, as the number of the transformer layer increases, the information retrieval metrics decrease. This observation demonstrates that more transformer layers are not beneficial. With the number of layers increasing, the candidate's original information will be lost, influencing the performance.

\begin{figure}[tb]
  \center{\includegraphics[width=0.48\textwidth, height=6cm] {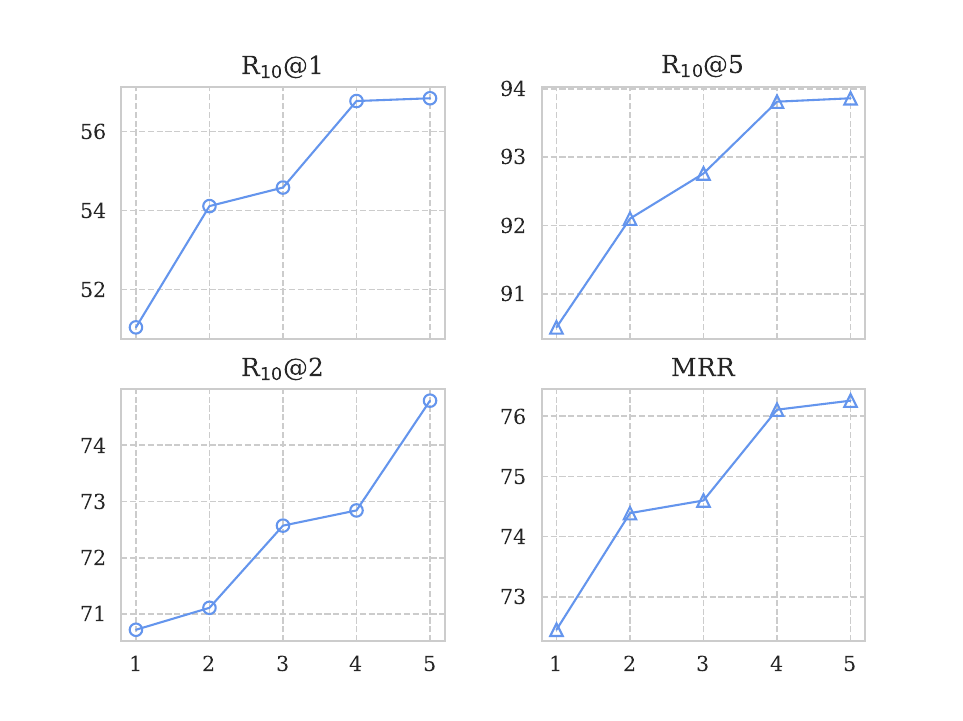}}
  \caption{The effect of fine-grained degree $k$.
  Except for the fine-grained degree $k$, other hyper-parameters keep the same.}
  \label{img:hyper_fg_k}
\end{figure}

\paragraph{Fine-grained Degree $k$} 
The hard negative selection is removed in this subsection, and only the positive instances expansion is tested.
As shown in Figure \ref{img:hyper_fg_k}, with the fine-grained degree $k$ increasing, more and more short and fine-grained positive training samples will be created.
It can be found that the augmented positive samples bring huge improvement. As $k$ increases, the performance becomes better.

\section{Inference Speed}
\label{sec:Inference Speed}
This section compares our proposed DR-BERT model, representative dialogue response selection models, and the dense retrieval model on inference efficiency. Specifically, we test their re-rank inference speed on the RRS corpus by providing the different sizes of the candidate set (10, 50, 100, and 1000 candidates, respectively).

We test the average re-rank inference speedup of following models on our proposed RRS corpus: SMN \cite{wu-etal-2017-sequential}, MSN \cite{Yuan2019MultihopSN}, SA-BERT \cite{Gu2020SpeakerAwareBF}, BERT-FP \cite{han-etal-2021-fine}, ColBERT \cite{Khattab2020ColBERTEA}, and our proposed DR-BERT. 
As shown in Table \ref{tab:rerank_inference_speed}, we can make the following conclusions:
(1) Compared with the cross-encoder models (SMN, MSN, SA-BERT, BERT-FP), dense retrieval models (ColBERT, DR-BERT) has the obvious superiority on the inference speed;
(2) The inference speed of the dense retrieval model (ColBERT, DR-BERT) could be improved further by re-using the pre-computed candidate representations;
(3) Although DR-BERT ranks slightly slower than the DR-BERT w/o. IL, DR-BERT still achieves comparable or faster inference speed than ColBERT. This observation demonstrates that the computational overhead of the interaction layer is limited and negligible;
(4) When the candidate set's size is small, such as 10 and 50, the DR-BERT achieves the comparable inference speed with the very efficient ColBERT model. With the size of the candidate set increasing, the DR-BERT ranks even faster than ColBERT. This observation demonstrates that our DR-BERT model is more efficient when the candidate set is bigger.


\begin{table}[tb]
\small
\renewcommand{\arraystretch}{1.2}
\setlength{\tabcolsep}{3.2pt}
        \begin{center}
            \scalebox{0.9}{
                \begin{tabular}{c|cccc}
                \hlinewd{0.75pt}
                \multirow{2}{*}{\textbf{Models}} & \multicolumn{4}{c}{\textbf{Re-rank Inference Speedup}}                                                                                 \\ \cline{2-5} & \multicolumn{1}{c|}{\textbf{RRS-10}}& \multicolumn{1}{c|}{\textbf{RRS-50}} & \multicolumn{1}{c|}{\textbf{RRS-100}} & \textbf{RRS-1000}\\ \hlinewd{0.75pt}
                \textbf{SMN}& \multicolumn{1}{c|}{1.0x}&\multicolumn{1}{c|}{1.0x}& \multicolumn{1}{c|}{1.0x}&1.0x              \\ 
                \textbf{MSN}&\multicolumn{1}{c|}{2.07x}&\multicolumn{1}{c|}{2.20x}&\multicolumn{1}{c|}{2.07x}&1.97x              \\ 
                \textbf{SA-BERT}&\multicolumn{1}{c|}{8.51x}&\multicolumn{1}{c|}{12.07x}& \multicolumn{1}{c|}{11.43x}&10.05x         \\ 
                \textbf{BERT-FP}& \multicolumn{1}{c|}{8.81x}&\multicolumn{1}{c|}{11.91x}& \multicolumn{1}{c|}{11.27x}&9.99x           \\ \hlinewd{0.25pt}
                \textbf{ColBERT w/o. cache}&\multicolumn{1}{c|}{5.27x}&\multicolumn{1}{c|}{19.87x}&\multicolumn{1}{c|}{21.72x}&20.74x \\ 
                \textbf{ColBERT}& \multicolumn{1}{c|}{11.58x}&\multicolumn{1}{c|}{52.46x}&\multicolumn{1}{c|}{76.84x}&217.66x
                \\ \hlinewd{0.25pt}
                \textbf{DR-BERT w/o. cache}&\multicolumn{1}{c|}{6.00x}&\multicolumn{1}{c|}{19.47x}& \multicolumn{1}{c|}{22.01x}&21.28x  \\
                 \tabincell{c}{\textbf{DR-BERT}}& \multicolumn{1}{c|}{10.48x}& \multicolumn{1}{c|}{48.26x}& \multicolumn{1}{c|}{84.79x}&489.68             \\
                 \textbf{DR-BERT w/o cache, IL}&\multicolumn{1}{c|}{6.08x}&\multicolumn{1}{c|}{20.63x}&\multicolumn{1}{c|}{22.73x}&21.52x   \\
                \textbf{DR-BERT w/o IL}&\multicolumn{1}{c|}{\textbf{13.06x}}&\multicolumn{1}{c|}{\textbf{63.53x}}&\multicolumn{1}{c|}{\textbf{108.14x}}&\textbf{502.67x} \\
                \hlinewd{0.75pt}
                \end{tabular}
            }
        \end{center}
    \caption{Comparison of re-rank inference speedup with different sizes of candidate set. The batch size for all of the models are the same. IL denotes the interaction layer.}
    \label{tab:rerank_inference_speed}
    \vspace{-0.8cm}
\end{table}

\section{Conclusions and Future Work}
In this work, we propose an effective and efficient dense retrieval solution, DR-BERT, for dialogue response selection. We introduce three simple yet effective training strategies to further augment DR-BERT's performance. For a comprehensive evaluation, we conduct extensive experiments on four benchmark datasets. The experimental results demonstrate the apparent superiority of DR-BERT over existing cross-encoder models. In future work, we will explore more training strategies to optimize the DR-BERT effectively.

\appendix

\section{RRS Ranking Experiment}


Human annotation is the most reliable metric, but it is costly, time-consuming, and irreproducible. The information retrieval metrics evaluate baselines automatically but may deviate from the human evaluation results. To address this dilemma, we design the ranking experiment for the RRS corpus, which could accurately and automatically measure the performance of dialogue response selection models.
Compared with the re-rank experiment, the ranking experiment contains much harder candidates, and their correlations with conversation context are carefully annotated by professional human annotators.

\subsection{RRS Ranking Test Set}
The following steps are adopted to build the RRS ranking test set that contains 800 dialogue sessions: (1) strong baselines generate 10 candidates for each context in the RRS test set, such as docTTTTTquery and DR-BERT; (2) 8 professional annotators are hired to rate these context-response pairs in terms of the quality on 1 to 5 (5 for the best);
(3) the average of their labeling scores is taken as the final correlation scores for candidates.

\subsection{Evaluation Protocol and Metrics}
Given the ranking scores and the accurate correlation scores that human annotators annotate, the Normalized Discounted Cumulative Gain at position k (NDCG@3 and NDCG@5) \cite{Jrvelin2002CumulatedGE} is used to measure the performance of the models.

\begin{table}[h]
\small
\renewcommand{\arraystretch}{1.2}
\setlength{\tabcolsep}{3.2pt}
  \begin{center}
    \scalebox{0.95}{
      \begin{tabular}{ccc}
      \hlinewd{0.75pt}
      \textbf{Models}         & \textbf{NDCG@3}   & \textbf{NDCG@5}  \\ \hlinewd{0.75pt}
      BERT                    & 0.625             &  0.714           \\
      BERT-FP                 & 0.609             &  0.709           \\
      SA-BERT$_{\rm BERT-FP}$ & 0.674             &  0.753           \\
      ColBERT$_{\rm BERT-FP}$ & 0.651             &  0.738  \\
      Poly-encoder            & 0.679             &  0.765           \\
      DR-BERT & \textbf{0.710}    & \textbf{0.783}           \\ \hlinewd{0.75pt}
      \end{tabular}
    }
    \caption{NDCG@3 and NDCG@5 scores.}
    \label{tab:ranking_quality_test_set}
  \end{center}
  \vspace{-0.8cm}
\end{table}

\subsection{Baselines}
We test the following baselines on the RRS ranking test set: BERT \cite{Whang2020AnED}, SA-BERT$_{\rm BERT-FP}$ \cite{Gu2020SpeakerAwareBF}, BERT-FP \cite{han-etal-2021-fine}, Poly-encoder \cite{Humeau2020PolyencodersAA}, ColBERT \cite{Khattab2020ColBERTEA,Santhanam2021ColBERTv2EA}, and DR-BERT. Note that the BERT-FP pretraining checkpoint is used for SA-BERT model (SA-BERT$_{\rm BERT-FP}$) and ColBERT (ColBERT$_{\rm BERT-FP}$).

\subsection{Experimental Results}
The experimental results on this RRS ranking test set are shown in Table \ref{tab:ranking_quality_test_set}. It can be found that the DR-BERT model achieves the best NDCG@3 and NDCG@5 scores on the RRS ranking test set. This observation suggests that the DR-BERT model
ranks the given candidates more accurately than other cross-encoder baselines, such as BERT-FP and SA-BERT$_{\rm BERT-FP}$.

\bibliographystyle{ACM-Reference-Format}
\bibliography{sample-base}

\end{document}